\documentclass[square,sort,comma,numbers]{article}
\usepackage[square,sort,comma,numbers]{natbib}

\usepackage{subcaption}

\usepackage{verbatim}

\usepackage{arxiv}
\usepackage{graphicx}

\usepackage[utf8]{inputenc} % allow utf-8 input
\usepackage[T1]{fontenc}    % use 8-bit T1 fonts
\usepackage{hyperref}       % hyperlinks
\usepackage{url}            % simple URL typesetting
\usepackage{booktabs}       % professional-quality tables
\usepackage{amsfonts}       % blackboard math symbols
\usepackage{nicefrac}       % compact symbols for 1/2, etc.
\usepackage{microtype}      % microtypography
\usepackage{lipsum}

\title{Improving Model-Based Control and Active Exploration with  Reconstruction Uncertainty Optimization}

\author{
  Norman Di Palo\thanks{Work done while intern at Curious AI.} \\
  Sapienza University of Rome
  \\
  Rome, Italy \\
  \texttt{normandipalo@gmail.com}
  \And 
  Harri Valpola \\
  Curious AI \\
  Helsinki, Finland
  }

\begin{document}
\maketitle

\begin{abstract}
Model-based predictions of future trajectories of a dynamical system often suffer from inaccuracies, forcing model-based control algorithms to re-plan often, thus being computationally expensive, sub-optimal and not reliable. In this work, we propose a model-agnostic method for estimating the uncertainty of a model's predictions based on reconstruction error, using it in control and exploration. As our experiments show, this uncertainty estimation can be used to improve control performance on a wide variety of environments by choosing predictions of which the model is confident. It can also be used for active learning to explore more efficiently the environment by planning for trajectories with high uncertainty, allowing faster model learning.
\end{abstract}

\section{Introduction}
\nocite{*}
Model-based Reinforcement Learning refers to a family of algorithms and methods that learn a model of a dynamical system (a "world model"), then use it to plan actions that optimize a particular cost or reward function. It is generally more sample efficient than model-free control, another family of methods that learn directly a policy, i.e. a mapping from states to actions, that optimizes a cost or reward. Still, a general issue with model-based control is caused by the inaccuracies that long predictions tend to show, due to accumulating errors. Furthermore, often the data distribution on which we test our predictions is different from the training distribution. This makes planning more difficult, since generally the trajectory generated by executing a series of actions is different from what expected. To deal with this, algorithms such as Model Predictive Control plan for a long horizon, but then executes only the first few actions of the plan, then re-plan. Re-planning often based on new observations helps controlling systems even in the presence of model inaccuracies, but this makes the algorithms significantly slower. \\ \\
In this work, we present a new architecture for controlling a system guided by the uncertainty estimation of the model predictions. While the model is learned using a recurrent neural network, a second feed forward network estimates the uncertainty of the predicted trajectory. The latter network is an autoencoder which takes a fixed number of time steps of states and actions $(s_t, a_t, \dots, s_{t+T}, a_{t+T})$ as input and tries to reconstruct them. The network, being trained on the same real-world samples seen during training by the model network, will reconstruct more accurately trajectories that are similar to what it has encountered during training, on which the model is thus more accurate, while its reconstruction error will be higher on unexplored areas of the states-actions space. This creates a dual concept of uncertainty/familiarity, where we can define a pattern as familiar if the network is able to efficiently compress and reconstruct it \cite{schmidhuber2008driven}. The intuition is similar to other works in the field of uncertainty estimation, that we discuss in Section 4. Nevertheless, while most works use an ensemble of models or stochastic dropout on a model to infer uncertainty with several feed-forward predictions, our method only needs to run one feed-forward prediction of a relatively small network. It is thus faster during prediction and also during backpropagation, used for gradient based optimization of the cost function with respect to the actions, and so more suited for real-time applications.  \\ \\
This new network can be used to select sequences of actions for which the model is more confident about its predictions. This leads to real trajectories that are more similar to the excepted ones, obtaining also a real reward that is closer to the one expected. Also, since the model is more confident about the outcome of the actions sequence, we can execute more actions before re-planning, thus having a faster algorithm for real-time applications. \\
Furthermore, the uncertainty estimation can be used to explore more rapidly and effectively the environment framing the problem of model learning as active learning. In active learning, an agent can choose the input samples that it wants to be labeled, with the goal of improving its performance rapidly. By choosing an action sequence that is predicted to have an uncertain outcome, we can plan for exploration, thus guiding the sample collection phase towards areas where the model is less accurate and can therefore learn more. This leads to faster model learning, since the algorithm rapidly explores most unvisited areas of the state-action space.
After visiting unexplored areas, both the world model and the uncertainty model will be trained on the new data. As a result, the accuracy of the world model on those trajectories will increase, while the uncertainty will decrease, making them quickly uninteresting for the agent to explore again. \\ \\
%Recent works in the area of uncertainty estimation [] often use an ensemble of models trained on different parts of the data or stochastic dropout on the same model to generate different outputs of the same inputs, estimating uncertainty as the degree of disagreement between the models. These methods are complementary to this work and could be used in conjunction to have a second uncertainty estimation. Nonetheless, we argue that having to run forward predictions with an ensemble of models (and then optimizing through gradient descent) is much slower than using a single feed forward network, and could be too computationally expensive in real-time applications. \\ \\
The paper is structured as follows: in Section 2 we will provide an introduction to reinforcement learning, model learning and model-based control, and will describe the network architectures and algorithms that we developed. In Section 3, we will show the results of our experiments focusing on control of different systems: an industrial chemical process and a quadcopter. These experiments show how our new method based on the Uncertainty Model improves the control performance on all the systems without needing more real-world data, lowering the error of predictions and obtaining an higher reward. Furthermore, we show the results of the experiments focusing on model learning: by planning actions that explore areas with high uncertainty, we can learn more efficiently a model that is accurate on a wider part of the state-action space. Experiments show how this technique led to better models on all the environments. In Section 4 we will discuss related work from the scientific literature. Section 5 concludes the paper with final considerations and further work. \\
The primary contributions of this work are the following: (1) we introduce a new method for predicting and optimizing uncertainty in model-based control based on reconstruction error of state-action sequences. (2) We empirically show that this method can be easily added to the usual optimization procedures of  model-based control to both achieve better control and more efficient exploration, independently of the world model. (3) We experimentally show its benefits on control and active learning on different environments: an industrial chemical process, a quadcopter and a quadruped robot.

\section{Architecture and Method}
\subsection{Preliminaries of Reinforcement Learning and Model-Based Control}
In Reinforcement Learning, at every time step $t$ an agent is in a state $s_t \subset S$ and receives an observation $o_t \subset O$. The agent can execute an action $a_t \subset A$, transitioning into a new state following the stochastic dynamical equation $ s_{t+1} \sim p( s_{t+1} | s_t, a_t) $, receiving a reward $ r_{t} = R(s_t, a_t)$. This is often formalized as a Markov Decision Problem (MDP), or a Partially Observable Markov Decision Problem (POMDP) if the state is not fully observable. In this work, though, we will refer to $s_t$ as the observed state of the agent, even if the full state is not observable.  The goal is to find a policy $a_t = \pi(s_t)$ that will bring the highest possible cumulative reward, $ \sum_{t=1}^T r(s_t, a_t)$ subject to $s_{t+1} \sim p( s_{t+1} | s_t, a_t)$. 
While in model-free reinforcement learning the agent usually learns a policy as a function mapping from observations to actions, in model-based reinforcement learning we select actions based on the prediction of future trajectories, using a learned model. We can plan ahead to generate a sequence of actions   $a_0, a_1, \dots , a_H = \arg \max \sum_{t=1}^H r(a_t, s_t) $ subject to $s_{t+1} = \hat{f}( s_t, a_t) $ , where $H$ denotes our planning horizon and $\hat{f}$ is our learned model, that maximize a reward or minimize a cost.  \\ \\
A model can be learned using completely off-policy data. An agent can explore the dynamics of its environment by executing random actions, and collecting tuples $(s_t, a_t, s_{t+1})$. A neural network is then trained on this data $D$ to learn the function $\Delta s_{t+1} = \hat{f}_{\theta}(s_t, a_t)$ by gradient descent, where $\theta$ indicates the trainable parameters of the network. We train the network to predict the difference between $s_{t+1}$ and $s_t$ instead of predicting directly $s_{t+1}$ because this led to better performance. Since some environments are partially observable or non-Markovian, we have to infer the current state of the system from the recent history of observations and actions. To do so, we adopt a Recurrent Neural Network, in particular an LSTM, to learn the mapping $\Delta s_{t+1} = \hat{f}_{\theta}(s_t, a_t, s_{t-1}, a_{t-1}, \dots, s_{t-T}, a_{t-T})$, where $T$ is the time window size of the network.\\
A widely used model-based control algorithm is Model Predictive Control. It consists of sampling a series of possible future actions sequences, predicting the cumulative reward of each based on the model, choosing the trajectory with the highest predicted reward and then executing only the first action. After that, we transition into a new state, obtain a new observation, and re-plan. This algorithm is quite effective and can be used to control various simulated robotic agents, as shown in \cite{nagabandi2017neural}, using a neural network model, since the constant re-planning attenuates model inaccuracies. Nevertheless, a drawback of the algorithm is its computational expensiveness: predicting numerous future trajectories can be computationally intensive, especially when this is done in real-time at every time step on a real robotic system.

\begin{figure}[t]
  \centering
  \includegraphics[width=0.8\linewidth]{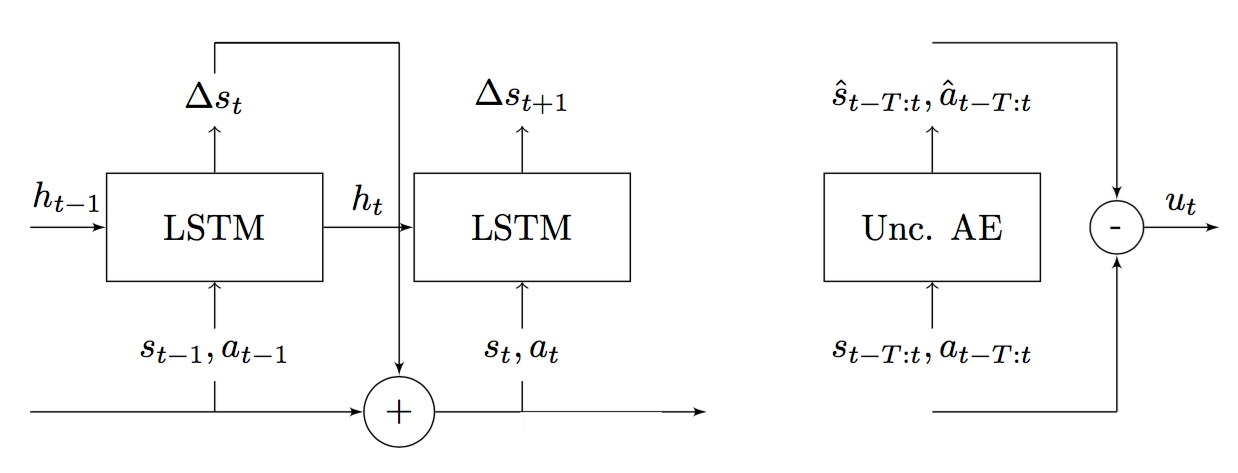}
  \caption{Overview of the architecture. Left: dynamical model, made of an LSTM that predicts $\Delta s_t$. In the prediction of a trajectory, the output $\Delta s_t$ is added to $s_{t-1}$ to have a prediction of $s_t$, that is fed as new input along with $a_t$. Right: uncertainty module, composed by an autoencoder. The uncertainty estimation is the mean squared error of input and output vectors (represented as - in the figure).}
  \label{fig:arch}
\end{figure}

\subsection{Uncertainty Estimation}
We propose an architecture for model-based control and model-based exploration that takes into account an estimated uncertainty of the trajectory predictions, based on reconstruction error of state-action trajectories. This method can help improving the control performance of a learned model, but also help learning more quickly and effectively a model of the world. Our architecture is composed of a feed forward network, independent from the dynamics model network, that is trained on the same state-action pairs $(s_t, a_t)$ collected during exploration. This network takes as input $T$ consecutive time steps of $(s_t, a_t)$, concatenated as one vector of dimension $1 \times (T + N + M)$, where $N, M$ are respectively the dimension of the state space and action space. The network is an auto-encoder, trained to output its input. It is trained by minimizing the mean squared error between its inputs and outputs using gradient descent
%\[(s_t, a_t, \dots, s_{t-T}, a_{t-T}) = AE_\phi (s_t, a_t, \dots, s_{t-T}, a_{t-T}) \]
\begin{equation}
\phi = argmin_\phi \frac{1}{T} \sum_{i=0}^{L-T} (\sum_{t = 0}^{T} (s_{i,t} - \hat{s}_{i,t})^2 + (a_{i,t} - \hat{a}_{i,t})^2)^{1/2}  
\end{equation}
where $\hat{s}_{t:t+T}, \hat{a}_{t:t+T} = AE_\phi (s_t, a_t, \dots, s_{t+T}, a_{t+T}) $
,  $\phi$ indicates the trainable parameters of the network and $L$ is the total time length of the collected data.  While a simple identity function would perfectly emulate this behaviour, the machine learning literature shows how neural networks often struggle to learn a general identity function outside their training data distribution \cite{he2016identity} \cite{trask2018neural}. The intuition is that the network will be able to reconstruct more accurately inputs similar to the ones it has encountered in its training set, collected via exploration, and thus trajectories on which our learned world model is more accurate. Trajectories out of this distribution will generate a higher reconstruction error. As observed in \cite{zhu2017deep}, this method can be seen as learning an embedding of the training trajectories, then at test time the reconstruction error gives an estimation of the distance between test samples and training samples in the embedding space. We can thus obtain a measure of the agent uncertainty on a particular time instant of the prediction by measuring the reconstruction error as the mean squared error between the vectors, 
%\[u_t = MSE((s_t, a_t, \dots, s_{t-T}, a_{t-T}) - AE_\phi (s_t, a_t, \dots, s_{t-T}, a_{t-T}))\] 
\begin{equation}
u_t = \frac{1}{T}(\sum_{i = t-T}^t (s_i - \hat{s}_i)^2 + (a_i - \hat{a}_i)^2)^{1/2} \end{equation} where $\hat{s}_{t-T:t}, \hat{a}_{t-T:t} = AE_\phi (s_t, a_t, \dots, s_{t-T}, a_{t-T}) $. Our architecture generates, for each time step, both a prediction of the future state $s_t$ and an estimated uncertainty of this prediction, $u_t$. 
\subsection{Model-Based Control with Uncertainty Minimization}
We now rephrase the problem of model-based control as an optimization process in the space of future actions that both maximizes a cumulative reward and minimizes a cumulative uncertainty 
\begin{equation} (a_t, a_{t+1}, \dots, a_{t+H}) =  arg max_{a_{t:t+H}} \sum _{i = t} ^{t+H} (\alpha r(s_i, a_i) - \beta u(s_{i:i-T}, a_{i:i-T}))  \end{equation},
where $\alpha, \beta$ are hyperparameters to weight the importance of having high cumulative reward or low cumulative uncertainty. This optimization procedure can either be solved by sampling $K$ random actions sequences, estimating the future trajectories and uncertainties and selecting the sequence with the highest cumulative value, or by gradient ascent, starting from an initial actions sequence and then optimizing the value with respect to the actions. The architecture, being composed of the sum of the outputs of two differentiable networks, is completely differentiable. We can then execute backpropagation through time to optimize the actions sequences. \\
As our experiments show (Section 3), this optimization procedure allows the agent to execute an higher number of actions before re-planning. The agent can execute actions until the estimated uncertainty has grown over a threshold, at which point it stops, receives the past observations and re-plan. This results in a faster and more reliable algorithm, able to execute even tens or hundreds of actions in open-loop.
\subsection{Model-Based Exploration with Uncertainty Maximization}
The uncertainty estimate can also be used during exploration to plan for actions sequences that are predicted to bring the system into unexplored state-action space areas. The procedure is very similar to the control phase, but now we optimize actions for high uncertainty, regardless of a cost or reward function. 
\begin{equation} (a_t, a_{t+1}, \dots, a_{t+H}) =  arg max_{a_{t:t+H}} \sum _{i = t} ^{t+H}  u(s_{i:i-T}, a_{i:i-T})  \end{equation}
We solve this optimization problem with the same techniques described before: we can either sample possible actions sequences and choose the one with the highest value, or optimize an initial action sequence through gradient ascent. In many real-world cases, we would like our agent to be curious, thus exploring areas with high uncertainty, but also cautious enough to avoid uncertainties above a certain threshold. In that case, we can simply stop the gradient ascent optimization phase or discard sequences that have an estimated uncertainty above a threshold.

\begin{figure}[t]
  \centering
  \begin{subfigure}[b]{0.4\linewidth}
    \includegraphics[width=\linewidth]{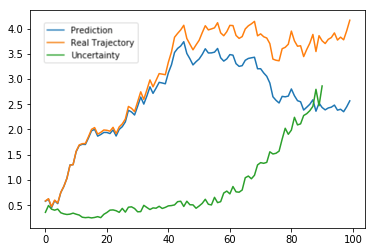}
  \end{subfigure}
  \begin{subfigure}[b]{0.4\linewidth}
    \includegraphics[width=\linewidth]{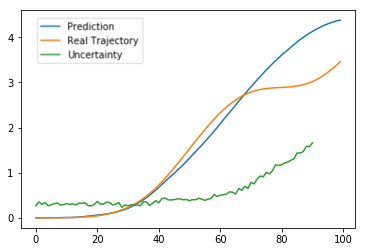}
  \end{subfigure}
  \caption{Examples of trajectory and uncertainty predictions, in the drone environment. Blue: predicted trajectory, orange: real trajectory, green: estimated uncertainty. Notice how the uncertainty rapidly grows when the error increases. This gives a good estimation of how many actions in the future we can execute before needing to re-plan, thus speeding up real-time execution considerably. The uncertainty is scaled to match the trajectories scale in order to be plotted.}
  \label{fig:coffee}
\end{figure}

\section{Experiments}
In this section we show and discuss the results of several experiments conducted on different environments. The goals of the experiments are first to empirically show that uncertainty-aware control yields better results, with a lower prediction error and an higher cumulative reward. Secondly, to show how uncertainty-guided exploration allows better model learning with the same number of samples with respect to random exploration. The environments that we used are the following: first, a recent simulator of an industrial chemical process, Industrial Benchmark \cite{hein2017benchmark}, designed, as the authors state, to benchmark control algorithms on a chemical process that simulates the difficulties of real-world scenarios, such as non-Markovian observations and changing dynamics. Second, a simulated and simplified drone/quadcopter, which doesn't include aerodynamics effects. Third, a simulated quadruped robot, the Minitaur, using the Bullet physics-engine. 

\begin{figure}[t]
  \centering
  \begin{subfigure}[b]{0.3\linewidth}
    \includegraphics[width=\linewidth]{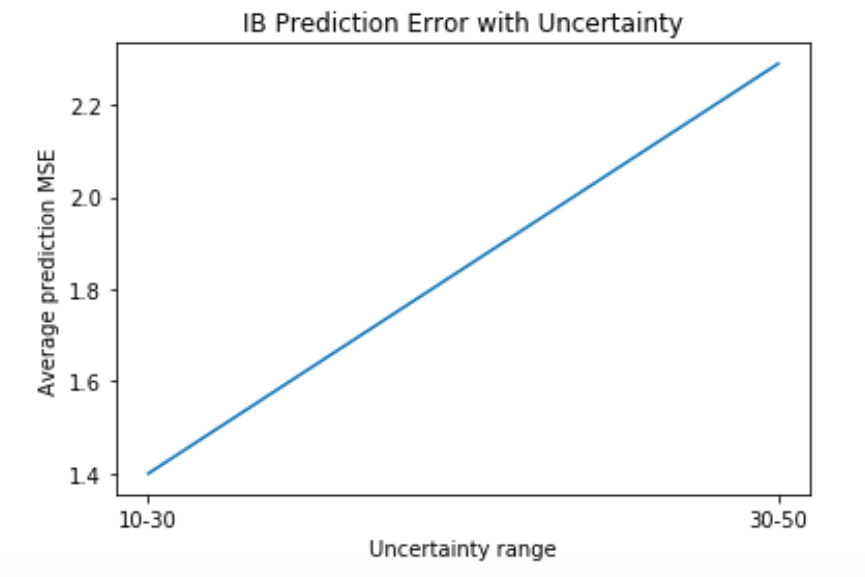}
  \end{subfigure}
  \begin{subfigure}[b]{0.3\linewidth}
    \includegraphics[width=\linewidth]{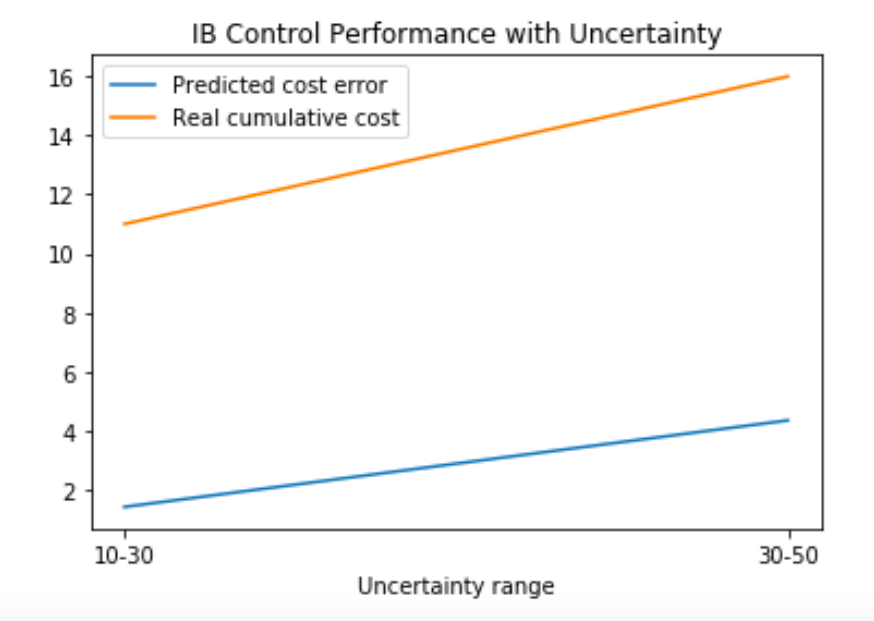}
  \end{subfigure}
  \begin{subfigure}[b]{0.3\linewidth}
    \includegraphics[width=\linewidth]{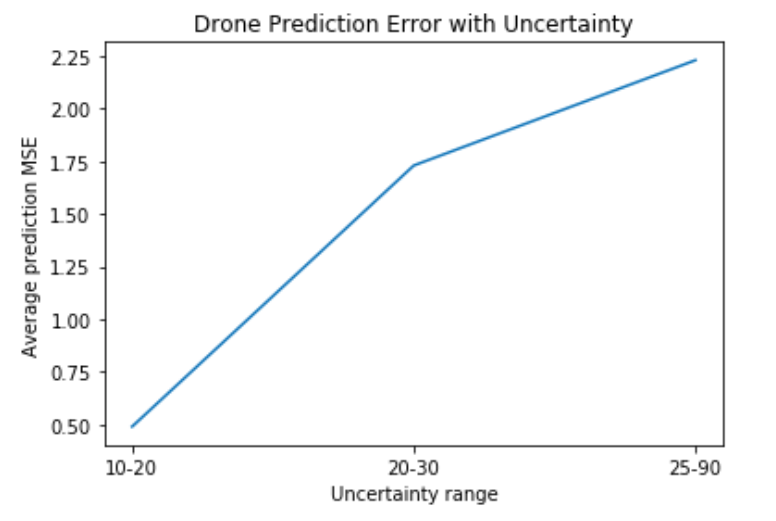}
  \end{subfigure}
  \begin{subfigure}[b]{0.3\linewidth}
    \includegraphics[width=\linewidth]{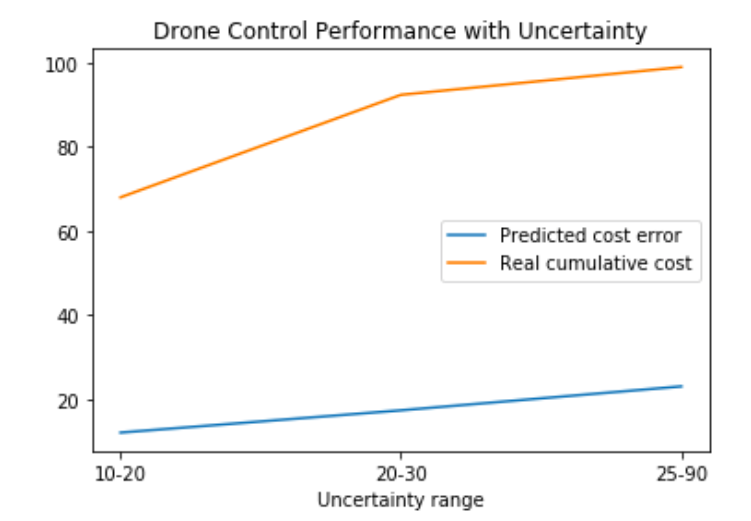}
  \end{subfigure}
  \begin{subfigure}[b]{0.3\linewidth}
    \includegraphics[width=\linewidth]{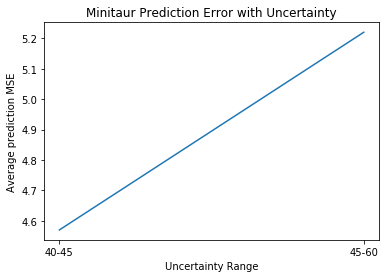}
  \end{subfigure}
  \begin{subfigure}[b]{0.3\linewidth}
    \includegraphics[width=\linewidth]{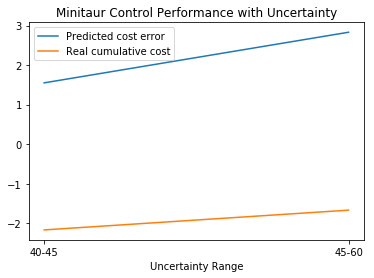}
  \end{subfigure}
  \caption{Prediction and control performance with varying uncertainty ranges for all the tested environments. Notice how, with lower predicted uncertainty, prediction error is lower, both for future trajectory and cost, and cumulative cost is also lower.}
  \label{fig:coffee}
\end{figure}

\subsection{Uncertainty-aware Model-based Control}
To benchmark our method, we compare it to the model-based approach proposed in \cite{nagabandi2017neural} on all the control problems. Thus, we first explore randomly the environment by sampling random actions and collecting observations in the form of $(s_t, a_t, s_{t+1})$. We fit a recurrent neural network to this data, as explained in Section 2, to predict the discrete change in state, $\Delta s_{t+1}$. Then, we plan a sequence of actions that optimize a cost function, by using both random sampling and gradient based optimization for different experiments. The two optimization methods gave similar results. We then execute the actions and compute an average mean squared error of the predicted trajectory and the real trajectory over 10 episodes. We also compute the predicted and real cumulative cost. Then, we repeat the experiments including the uncertainty prediction. We fit an autoencoder, as discussed in Section 2, to the same data gathered during random exploration. We then generate a plan by optimizing both for low cumulative cost and low uncertainty, as in Equation 3. We then compare the prediction errors and real cumulative costs of these two methods. 
In all of our experiments, we use as world model an LSTM with 32 units, followed by a hidden densely connected layer and a linear output layer. Our uncertainty network is composed of a single hidden layer with 100 units, thus being relatively small and allowing fast feed-forward inference and backpropagation of derivatives to optimize actions. The exploration phase is composed of 10 episodes of 500 time steps, for a total of 5000 samples. The planning horizon $H$ is 200, and the autoencoder time window $T$ is 10. We start by sampling 10 random actions after resetting the environments.\\
\textbf{Industrial Benchmark}: The Industrial Benchmark (IB) environment has 3 actions and 6 observations. It's non-Markovian, and the current state must be inferred by past observations and actions. It also has changing dynamics based on the recent history as explained in the paper \cite{hein2017benchmark}. 
%In Figure 3 we compare the prediction errors of the world model alone and the uncertainty-aware model, obtained by planning actions with gradient based optimization of Equation 3, using cost and uncertainty weights of $\alpha = 1, \beta = 0.2$ and then $\alpha = 1, \beta = 0.1$.
In Figure 3 we compare the prediction errors of the world model alone and the uncertainty-aware model. We averaged the results of MPC-based sampling of actions, choosing the lowest estimated cumulative cost among the sampled sequences of actions which also have an estimated cumulative uncertainty between 15 and 30 first, and then between 30 and 50, to showcase the results of trajectories with different ranges of predicted cumulative uncertainty. We ran 10 episodes and averaged the prediction error, computed as mean squared error of real and predicted states trajectory, and the difference between real and expected cumulative cost. In Figure 3 we show the average results of prediction MSE and error in cumulative cost estimation. As the experiments show, an lower uncertainty range, or equally an higher uncertainty weight in (3), brings lower prediction error, cumulative cost prediction error and generally lower real cumulative cost. Thus, the new optimization method yields better performance with the same number of samples. We obtained similar results using gradient-based optimization of actions, optimizing Equation 3 with $\alpha = 1$ and $\beta$ between $0.1$ and $0.2$.\\
\textbf{Drone}: The drone environment has 4 actions, the velocities of each propeller, and 12 observations: the world-frame position and velocities, angles and angular velocities expressed in Euler angles. The environment is thus Markovian, but we still use the same recurrent world model to benchmark the same model. The cost function is computed as the absolute distance of the drone position in world coordinates and the goal position $(1,1,1)$. In these experiments, we sample sequences of actions until we find a predicted cumulative uncertainty that is between a certain range and predicted cumulative cost under a threshold. The cost threshold is fixed, but we use different uncertainty ranges: $10 - 20, 20 - 30, 25 - 90$. As Figure 3 shows, the mean squared error of predictions and real trajectories grows as the uncertainty range grows. Similarly, the predicted cost error grows as well. The new optimization procedure yields better performance in this case as well. \\
\textbf{Minitaur}: The Minitaur is a quadruped robot developed by Ghost Robotics. In this work, we use a simulated version of the robot using Bullet physics engine. The environment has 8 actions, representing the torque of the leg motors. The cost function is the negative squared distance travelled along the $x$ direction, hence the optimization algorithm will push the robot forward. As before, we actuate the actions with the lowest predicted cost, and with a predicted uncertainty inside a certain range. Also in this case, we show in our experiments (Figure 3) how the mean squared error of predictions and real trajectories grows as uncertainty estimate grows. At the same time, the error between predicted and actual cost grows when the uncertainty estimate grows.

\begin{comment}

\begin{figure}[t]
  \centering
  \begin{subfigure}[b]{0.3\linewidth}
    \includegraphics[width=\linewidth]{expl_ib.png}
  \end{subfigure}
  \begin{subfigure}[b]{0.3\linewidth}
    \includegraphics[width=\linewidth]{expl_drone.png}
  \end{subfigure}
  \begin{subfigure}[b]{0.3\linewidth}
    \includegraphics[width=\linewidth]{expl_mini.png}
  \end{subfigure}
  \caption{Mean Squared Error of trajectory predictions on the validation set (100 trajectories of 100 steps): the plot shows the mean and standard deviation. Same model trained on uncertainty-guided exploration data and random exploration data. Average of 5 experiments, each with 10 re-trainings of the model on the same data to alleviate random initialization effects.}
  \label{fig:coffee}
\end{figure}
\end{comment}

\begin{table}
\begin{center}
    \begin{tabular}{| l | l | l |}
    \hline
    Env. & Random Explor. & Active Unc. Explor.  \\ \hline
    \textbf{IB} & $\mu$ = 44.8, $\sigma$ = 8.1 & \textbf{$\mu$ = 26.2, $\sigma$ = 7.1}  \\ \hline
    \textbf{Drone} & $\mu$ = 10.3 , $\sigma$ = 1.8 & \textbf{$\mu$ = 7.6, $\sigma$ = 1.4}  \\ \hline
    \textbf{Minitaur} & $\mu$ = 7.5, $\sigma$ = 2.4 & \textbf{$\mu$ = 6.4, $\sigma$ = 1.3}  \\
    \hline

    \end{tabular} 
    \caption{Mean Squared Error of trajectory predictions on the validation set (100 trajectories of 100 steps). Same model trained on uncertainty-guided exploration data and random exploration data. Average of 5 experiments, each with 10 re-trainings of the model on the same data to average random initialization effects.}
\end{center}
\end{table}

\subsection{Uncertainty-Guided Active Exploration and Model Learning}
In these experiments, we benchmark empirically if uncertainty-guided active exploration can help building better models with the same amount of samples. The experiments are structured as follows: we start with an initial phase of random exploration of 1e4 time steps. We train the world model and uncertainty model, which have the same architecture as before, on the collected data. We then sample 100 trajectories as a validation set, on which we later benchmark the prediction error of the model. We proceed to a second phase of exploration. In one case, we plan sequences of actions for which the predicted cumulative uncertainty is higher than the average cumulative uncertainty of a certain threshold (in the range of 1.2 - 1.5 times higher). We empirically showed, in the previous series of experiments, that higher uncertainty strongly correlates with higher prediction error, mostly due to executing trajectories that are far from the training distribution. The intuition is that collecting several trajectories with higher prediction error helps building a better model in a larger part of the state-action space, instead of exploring the same areas already explored on which the model has low error. We gather 1.5e4 time steps with uncertainty-guided exploration, as 150 episodes of 100 steps. We also gather 1.5e4 time steps with random exploration. We finally retrain the world model first on the union of the initial 1e4 time steps and new 1.5e4 time steps gathered by planning for high uncertainty, then with the initial data and newly randomly collected data, obtaining two datasets of 2.5e4 time steps. After training, we average the prediction error on the validation set of 100 trajectories. Since random initialization of the network can impact the benchmark performance, we average the results over 10 different retraining. Furthermore, we repeat the experiments from the beginning 5 times for each environment and average the results. \\
In Table 1 we show the results of the experiments on the IB, Drone and Minitaur environments. As the experiments show, the model learned with uncertainty-guided active exploration is consistently more accurate of the model learned with random exploration. This leads to better control performance as well, since prediction accuracy is crucial in model-based control.

\section{Related Work}
The literature of Deep Reinforcement Learning has rapidly grown in recent years. Many recent papers have focused on new algorithms for efficient exploration, model-based control and uncertainty estimation.  \\\\
\textbf{Exploration:} An agent has to first explore its unknown environment before being able to learn a policy that generates high cumulative reward. Exploration can be conducted by learning a model-free policy, as in the case of \cite{pathak2017curiosity}, where an agent learns a policy that maximizes the error in prediction of an inner world model. Similarly, the recent work \cite{burda2018exploration} learns a model-free policy that maximizes the error between a randomly initialized network and a second network that tries to mimic its outputs, with input the current state. The intuition behind this method is similar to ours: as the authors state, the error will be higher in states that are rarely visited. In our work we use an autoencoder for computing the reconstruction error instead of a randomly initialized network.  Furthermore, we use a completely model-based approach, by planning for exploration, instead of learning a model-free policy that is generally less sample efficient. Another recent work \cite{shyam2018model} learns an ensemble of models and plans to explore trajectories where the model disagrees. Again, the models will mostly disagree on states that are rarely visited. In our work, we use a single uncertainty network instead of an ensemble of models, since using a single feed forward network for uncertainty prediction is faster than computing the outputs of several networks. Furthermore, we show that uncertainty can also be used to improve control. The recent work \cite{lowrey2018plan} uses an ensemble of model as well to predict the value (sum of future cumulative discounted rewards) for each state, and explores state where the models disagree (optimism in the face of uncertainty, as stated by the authors). An older work \cite{Schmidhuber91adaptiveconfidence} studies the general structure of an agent with adaptive curiosity and confidence, able to explore areas with low-confidence to improve its world model. Similarly to this framework, our agent explores, driven by curiosity, areas where it predicts high uncertainty, but then, training its uncertainty network on the new data, it will no longer find those trajectories interesting. Its world model will improve in those areas and the uncertainty will decrease, thus showing an adaptive curiosity and confidence.\\ \\
\textbf{Uncertainty:} The recent literature of deep learning has proposed several methods to evaluate the uncertainty of the prediction of a deep neural network. Several works propose the use of Bayesian Neural Networks, or approximations of these architectures \cite{gal2016uncertainty}. Many papers, as discussed, approximate a Bayesian posterior by either using an ensemble of models, or applying stochastic dropout to a network and computing several outputs for the same input, and computing then mean and variance. An interesting recent work \cite{zhu2017deep} uses both stochastic dropout and a recurrent autoencoder to estimate the uncertainty of a time series estimation. The authors feed the cell states of the encoder LSTM directly in the prediction network, while we use the reconstruction error of a feed forward network as an uncertainty estimation. Furthermore, we also optimize
actions through the model for control or exploration. The paper \cite{schmidhuber2008driven} elaborates on the connection of compression and understanding, and the various interpretation of curiosity, discovery and familiarity.

\textbf{Model-based Control:} The work \cite{nagabandi2017neural} shows how a neural network can learn a world model to control agents simulated in the MuJoCo physics-engine through step-by-step re-planning using Model Predictive Control. Similarly, the work \cite{lowrey2018plan} showed how it is possible to improve model-based control by adding a value function approximator. This method is complementary to ours and could be integrated into the general architecture as future work. The paper \cite{kahn2017uncertainty} adopts an uncertainty-aware model-based control technique to autonomously fly a drone near obstacles, using dropout to estimate uncertainty. \\\\

\section{Conclusion}
In this work, we presented a new architecture and method for estimating the uncertainty of world model's trajectory predictions based on reconstruction error, and for optimizing it to both obtain better control performance and more efficient exploration and model learning. The uncertainty network, being a single and relatively small feed-forward network, is generally fast to run and suited to real-time applications with respect to ensemble of models used in the recent literature. This method is also model-agnostic: it can be easily added to any chosen world-model, obtaining virtually the same estimations of familiarity and uncertainty as reconstruction error. Furthermore, the architecture designs of the world model and uncertainty model are independent, allowing for more flexibility. \\ The ability to obtain more reliable predictions of the future allows agents to execute a higher number of actions in open-loop before the need to re-plan, thus making the general algorithm faster. Furthermore, a more efficient exploration allows real robots to learn a better model using less real-world data, that is often the real bottleneck of real world deep reinforcement learning. We argue that model-based active exploration can help improve data-efficiency for various control tasks, as showed by our experiments.

\section*{Acknowledgments}
We wish to thank Antti Rasmus, Rinu Boney, Teemu Tiinanen, Mathias Berglund, Jussi Sainio and Jari Rosti for the helpful discussions, technical support and sharing of ideas, and all the people at Curious AI for their contribution in creating a stimulating and enjoyable workplace.

\bibliographystyle{plainnat} % We choose the "plain" reference style

\begin{thebibliography}{10}

\bibitem{burda2018exploration}
Yuri Burda, Harrison Edwards, Amos Storkey, and Oleg Klimov.
\newblock Exploration by random network distillation.
\newblock {\em arXiv preprint arXiv:1810.12894}, 2018.

\bibitem{erez2013integrated}
Tom Erez, Kendall Lowrey, Yuval Tassa, Vikash Kumar, Svetoslav Kolev, and
  Emanuel Todorov.
\newblock An integrated system for real-time model predictive control of
  humanoid robots.
\newblock In {\em Humanoid Robots (Humanoids), 2013 13th IEEE-RAS International
  Conference on}, pages 292--299. IEEE, 2013.

\bibitem{gal2016uncertainty}
Yarin Gal.
\newblock Uncertainty in deep learning.
\newblock {\em University of Cambridge}, 2016.

\bibitem{he2016identity}
Kaiming He, Xiangyu Zhang, Shaoqing Ren, and Jian Sun.
\newblock Identity mappings in deep residual networks.
\newblock In {\em European conference on computer vision}, pages 630--645.
  Springer, 2016.

\bibitem{hein2017benchmark}
Daniel Hein, Stefan Depeweg, Michel Tokic, Steffen Udluft, Alexander Hentschel,
  Thomas~A Runkler, and Volkmar Sterzing.
\newblock A benchmark environment motivated by industrial control problems.
\newblock In {\em Computational Intelligence (SSCI), 2017 IEEE Symposium Series
  on}, pages 1--8. IEEE, 2017.

\bibitem{hochreiter1997long}
Sepp Hochreiter and J{\"u}rgen Schmidhuber.
\newblock Long short-term memory.
\newblock {\em Neural computation}, 9(8):1735--1780, 1997.

\bibitem{kahn2017uncertainty}
Gregory Kahn, Adam Villaflor, Vitchyr Pong, Pieter Abbeel, and Sergey Levine.
\newblock Uncertainty-aware reinforcement learning for collision avoidance.
\newblock {\em arXiv preprint arXiv:1702.01182}, 2017.

\bibitem{lowrey2018plan}
Kendall Lowrey, Aravind Rajeswaran, Sham Kakade, Emanuel Todorov, and Igor
  Mordatch.
\newblock Plan online, learn offline: Efficient learning and exploration via
  model-based control.
\newblock {\em arXiv preprint arXiv:1811.01848}, 2018.

\bibitem{nagabandi2017neural}
Anusha Nagabandi, Gregory Kahn, Ronald~S Fearing, and Sergey Levine.
\newblock Neural network dynamics for model-based deep reinforcement learning
  with model-free fine-tuning.
\newblock {\em arXiv preprint arXiv:1708.02596}, 2017.

\bibitem{pathak2017curiosity}
Deepak Pathak, Pulkit Agrawal, Alexei~A Efros, and Trevor Darrell.
\newblock Curiosity-driven exploration by self-supervised prediction.
\newblock In {\em International Conference on Machine Learning (ICML)}, volume
  2017, 2017.

\bibitem{schmidhuber2008driven}
J{\"u}rgen Schmidhuber.
\newblock Driven by compression progress: A simple principle explains essential
  aspects of subjective beauty, novelty, surprise, interestingness, attention,
  curiosity, creativity, art, science, music, jokes.
\newblock In {\em Workshop on Anticipatory Behavior in Adaptive Learning
  Systems}, pages 48--76. Springer, 2008.

\bibitem{Schmidhuber91adaptiveconfidence}
Jürgen Schmidhuber.
\newblock Adaptive confidence and adaptive curiosity.
\newblock Technical report, Institut fur Informatik, Technische Universitat
  Munchen, Arcisstr. 21, 800 Munchen 2, 1991.

\bibitem{shyam2018model}
Pranav Shyam, Wojciech Ja{\'s}kowski, and Faustino Gomez.
\newblock Model-based active exploration.
\newblock {\em arXiv preprint arXiv:1810.12162}, 2018.

\bibitem{thrun1992active}
Sebastian~B Thrun and Knut M{\"o}ller.
\newblock Active exploration in dynamic environments.
\newblock In {\em Advances in neural information processing systems}, pages
  531--538, 1992.

\bibitem{trask2018neural}
Andrew Trask, Felix Hill, Scott~E Reed, Jack Rae, Chris Dyer, and Phil Blunsom.
\newblock Neural arithmetic logic units.
\newblock In {\em Advances in Neural Information Processing Systems}, pages
  8045--8054, 2018.

\bibitem{zhu2017deep}
Lingxue Zhu and Nikolay Laptev.
\newblock Deep and confident prediction for time series at uber.
\newblock In {\em Data Mining Workshops (ICDMW), 2017 IEEE International
  Conference on}, pages 103--110. IEEE, 2017.

\end{thebibliography}

\end{document}